\documentclass[journal]{IEEEtran}
\ifCLASSINFOpdf
\else
\fi
\usepackage{graphicx}
\usepackage{multirow}
\usepackage{epsfig}
\usepackage{graphicx}
\usepackage{amsmath}
\usepackage{amssymb}
\usepackage{amsthm}
\usepackage{bm}
\usepackage{mathtools}
\usepackage{algorithm,algcompatible,lipsum}
\usepackage{verbatim}
\usepackage{cite} 
\usepackage{flushend}

\usepackage{array}
\newcolumntype{P}[1]{>{\centering\arraybackslash}p{#1}}

\usepackage{subcaption}

\usepackage[pangram]{blindtext}

\newcommand{\dimin}{N}
\newcommand{\dimout}{M}

\def\L{{\mathcal{L}}}   
\def\X{{X}}   
\def\Y{{Y}}   
\def\R{\mathbb{R}}

\begin{document}
%
\title{Adaptive Weight Decay for Deep Neural Networks}
%

\markboth{}
{Shell \MakeLowercase{\textit{K. Nakamura, and B.-W. Hong}}: Adaptive Weight Decay for Deep Neural Networks}
%

%
%

\author{Kensuke~Nakamura,
        and~Byung-Woo~Hong*
\thanks{* corresponding author: B.-W. Hong e-mail: hong@cau.ac.kr }
\thanks{K Nakamura and B.-W. Hong are with the Computer Science Department, Chung-Ang University, Korea  (see https://www.image.cau.ac.kr/).}
}

%
%



\maketitle

\begin{abstract}
Regularization in the optimization of deep neural networks is often critical to avoid undesirable over-fitting leading to better generalization of model. One of the most popular regularization algorithms is to impose $L_2$ penalty on the model parameters resulting in the decay of parameters, called weight-decay, and the decay rate is generally constant to all the model parameters in the course of optimization. In contrast to the previous approach based on the constant rate of weight-decay, we propose to consider the residual that measures dissimilarity between the current state of model and observations in the determination of the weight-decay for each parameter in an adaptive way, called adaptive weight-decay (AdaDecay) where the gradient norms are normalized within each layer and the degree of regularization for each parameter is determined in proportional to the magnitude of its gradient using the sigmoid function. We empirically demonstrate the effectiveness of AdaDecay in comparison to the state-of-the-art optimization algorithms using popular benchmark datasets: MNIST, Fashion-MNIST, and CIFAR-10 with conventional neural network models ranging from shallow to deep. The quantitative evaluation of our proposed algorithm indicates that AdaDecay improves generalization leading to better accuracy across all the datasets and models.
\end{abstract}

\begin{IEEEkeywords}
Adaptive Regularization, Deep Learning, Neural Networks, Stochastic Gradient Descent, Weight-decay
\end{IEEEkeywords}

%
\IEEEpeerreviewmaketitle

%
%
%
\section{Introduction} \label{sec:intro}
\vspace{16pt}
The deep neural network model consists of a nested architecture of layers where the number of parameters is in millions\cite{bottou2010large,simonyan2014very,szegedy2015going,he2016deep,he2016identity,huang2017densely,bottou2018optimization}.
Due to its high degrees of freedom, the deep model can approximate linear and nonlinear functions; however it is always at risk of over-fitting to training data.
Thus the deep neural network requires regularization techniques in training process in order to achieve generalization, resulting in a good prediction for unknown data.
\par
Since the number of example data is also huge, the deep model is trained using the stochastic gradient descent (SGD)~\cite{robbins1951stochastic,rumelhart1988learning,zhang2004solving,bottou2010large,bottou2018optimization} 
that updates parameters using a small subset of data (mini-batch) in combination with regularization techniques.
A simple yet major regularization technique is, so-called, early stopping~\cite{prechelt1998early,zhang2016understanding} where the training process is terminated manually at a certain epoch before the validation loss increases.
The noise injection techniques, 
e.g.,  the mini-batch procedure that induces noise to gradients~\cite{zhu2019anisotropic} and the dropout that randomly zeros the activation of nodes~\cite{srivastava2014dropout},
give implicit regularization effects for the training process.
The weight-decay is an explicit regularization such that an additional penalty term is defined in the energy function and the regularization effect can be tuned by its coefficient.
%
In contrast to the recent development of adaptive methods on deep optimization, a constant weight-decay has been employed
while it played an important role in both classic and modern deep neural networks~\cite{krogh1992simple,simonyan2014very,chollet2017xception,zhang2016understanding}.
Layer-wise weight-decay was considered in~\cite{ishii2017layer,bengio2012practical} but it is limited to classical neural networks without the skip-connection.
\par
We propose an adaptive regularization method of weight-decay, called Adaptive Weight-Decay (AdaDecay), that varies in spatio-temporal domains during the training process and is beneficial to both shallow and deep neural network models.
The proposed AdaDecay determines the weight-decay rate in parameter-wise at each optimization iteration using the magnitude of the loss gradient 
that measures the difference between the current state of the model with the mini-batch data.
We normalize the gradient norm in each layer in order to make the algorithm independent to the model architecture and robust to hyper-parameter selection.
We also present experimental results of the presented AdaDecay in comparison to the state-of-the-art optimization algorithms using major benchmark datasets on the image classification with shallow and deep networks, where our AdaDecay has overtakes the others in validation accuracy.
\par
In the remaining of this paper we summarize the related studies with our contributions in Section~\ref{sec:works}
and then describe notations in Section~\ref{sec:prelimminary}. We present AdaDecay in Section~\ref{sec:adadecay}, followed by experimental results in Section~\ref{sec:experimental} and conclusion in Section~\ref{sec:conclusion}
%
%
%
%

%
%
%
\section{Related Work} \label{sec:works}

\vspace{3pt}
\noindent {\bf Learning Rate Annealing:}
The stochastic gradient, calculated by a subset of data, gives a noise to gradient and provides an implicit regularization effect~\cite{zhu2019anisotropic}.
In SGD, parameters are updated by subtracting the gradient with the stochastic noise multiplied by the learning rate.
The learning rate should shrink in order to reduce the noise and converge the algorithm.
To this aim, a variety of learning rate annealing, e.g. exponential~\cite{george2006adaptive} and staircase~\cite{smith2017don}, and the adaptive learning rates, e.g., AdaGrad~\cite{duchi2011adaptive}, have been proposed, 
The sophisticated adaptive techniques, e.g., RMSprop~\cite{tieleman2012lecture} and Adam~\cite{kingma2014adam}, enable parameter-wise control of the learning rates.
The drawback of learning rate techniques on the regularization is that it reduces or increases both the step-size and the noise.

\vspace{3pt}
\noindent {\bf Dropout} is another regularization technique that is in particular used with classical shallow networks.
The dropout zeros the activation of randomly selected nodes with a certain probability during the training process~\cite{srivastava2014dropout}.
The dropping rate is generally set to be constant but its variants have been considered with adaptive rates depending on parameter value~\cite{ba2013adaptive}, estimated gradient variance~\cite{kingma2015variational}, biased gradient estimator~\cite{srinivas2016generalized}, layer depth~\cite{huang2016deep}, or marginal likelihood over noises~\cite{noh2017regularizing}.
However, in fact, recent deep models do not support the dropout and its variants.  The reason may be that the number of parameters in a layer is relatively smaller than the the classic neural networks, and random masking to nodes can be erroneous to the model.

\vspace{3pt}
\noindent {\bf Energy Landscape:}
The geometrical property of energy surface is helpful in optimization of highly complex non-convex problems associated with deep network architecture.
It is preferred to drive a solution toward local minima on a flat energy surface that is considered to yield better generalization~\cite{hochreiter1997flat,Chaudhari2017EntropySGD,dinh2017sharp} where flatness is defined around the minimum by its connected region, its curvature of the second order structure, and the width of its basin, respectively.
A geometry-driven optimization based on SGD has been developed in deep learning problems such as Entropy-SGD~\cite{Chaudhari2017EntropySGD}.
In our approach, we do not attempt to measure geometric property of loss landscape such as flatness with extra computational cost, but instead consider explicit regularization to model parameters.

\vspace{3pt}
\noindent {\bf Variance Reduction:} 
The variance of stochastic gradients is detrimental to SGD, motivating variance reduction techniques~\cite{Roux2012,johnson2013accelerating,Chatterji2018OnTT,zhong2014fast,shen2016adaptive,Difan2018svrHMM,Zhou2019ASim} that aim to reduce the variance incurred due to their stochastic process of estimation, and improve the convergence rate mainly for convex optimization while some are extended to non-convex problems~\cite{allen2016variance,huo2017asynchronous,liu2018zeroth}. 
One of the most practical algorithms for better convergence rates includes momentum~\cite{sutton1986two}, modified momentum for accelerated gradient~\cite{nesterov1983method}, and stochastic estimation of accelerated gradient (Accelerated-SGD)~\cite{Kidambi2018Acc}.
These algorithms are more focused on the efficiency in convergence than the generalization of model for accuracy.

\vspace{3pt}
\noindent {\bf Weight-Decay:} is an explicit way of regularization such that a regularization term is added into the energy function. 
Specifically $L^2$-norm is used as the regularization term in order to penalize large weight values.
Different with the other implicit methods, e.g., stochastic update and dropout, one can directly control the regularization effect by the weight-decay coefficient.
The weight-decay coefficient is tuned by hand~\cite{simonyan2014very,chollet2017xception}, or learned by Bayesian optimization~\cite{snoek2015scalable,shahriari2016unbounded}.
However, in contrast to recent development of adaptive methods of dropout~\cite{ba2013adaptive,kingma2015variational,srinivas2016generalized,huang2016deep,noh2017regularizing} 
and learning-rate~\cite{duchi2011adaptive,tieleman2012lecture,kingma2014adam} in deep optimization, a constant weight-decay coefficient has been employed in usual.
Layer-wise weight-decay has been considered in~\cite{ishii2017layer,bengio2012practical} where different weight-decay coefficients are given for different layers of network model using the variance of gradients in layer.
The drawback of the layer-wise method~\cite{ishii2017layer,bengio2012practical} is that it assumes that layers are aligned in a single sequence.  
The skip-connection~\cite{he2016deep,he2016identity,Balduzzi2017shattered}, that is one of the key architectures in the recent deep networks, makes it non-trivial.
%
%

The main contributions of this work are three folds: 
First, we propose an adaptive regularization method (AdaDecay) in which parameter-wise weight-decay varies in spatio-temporal domains reflecting the currant state of model and the mini-batch data.
Second, the proposed AdaDecay determines the weight-decay rate of each parameter based on the norm of gradient normalized in layer.
This makes the algorithm independent to model architecture and beneficial to both shallow and deep neural network models.
Third, we empirically demonstrate the robustness and effectiveness of the presented AdaDecay in comparison to the state-of-the-art optimization algorithm using both shallow neural network models and modern deep models with three of the major benchmark datasets on image classification.
%

%
%
\section{Preliminary} \label{sec:prelimminary}
We consider an energy optimization problem based on a given set of training data in a supervised machine learning framework. Let $\chi = \{ (x_i, y_i) \}_{i = 1}^n$ be a set of training data where $x_i \in \X \subset \mathbb{R}^\dimin$ is the $i$-th input and $y_i \in \Y \subset \mathbb{R}^\dimout$ is its desired output. 
Let $h_w \colon \X \to \Y$ be a prediction function that is associated with its model parameters $w = ( w_1, w_2, \cdots, w_m ) \in \R^m$ where $m$ denotes the dimension of the feature space.
The objective of the supervised learning problem under consideration is to find an optimal set of parameters $w^*$ by minimizing the empirical loss $\L(w)$ that typically consists of a data fidelity term $\rho(w)$ and a regularization term $\gamma(w)$ as follows:
%
    \begin{align}
    w^* = \arg\min_w \mathcal{L}(w), \quad
    \mathcal{L}(w) = \rho( w ) + \lambda \, \gamma( w ), \label{eq:loss}
    \end{align}
%
where $\lambda > 0$ is a control parameter, called weight-decay coefficient, that determines the trade-off between the data fidelity term $\rho(w)$ and the regularization $\gamma(w)$.
The data fidelity term $\rho(w)$ is of the additive form over a set of training data $\{ (x_i, y_i) \}_{i=1}^n$ as follows:
%
    \begin{align}
    \rho(w) &= \frac{1}{n} \sum_{i=1}^n f_i( w ), \label{eq:fidelity}
    \end{align}
%
where $f_i( w )$ denotes a data fidelity incurred by a set of model parameters $w$ for a sample pair $(x_i, y_i)$.  
The data fidelity $f_i( w )$ is designed to measure the discrepancy between the prediction $h_w( x_i )$ with an input $x_i$ and its desired output $y_i$ for a given sample pair $(x_i, y_i)$.
The regularization $\gamma( w )$ is designed to impose a smoothness constraint to the solution space, thus avoid undesirable over-fitting to the model. 
The weight-decay coefficient $\lambda \in \R$ is determined based on the relation between the underlying distribution of the data and the prior distribution of the model. 
In the optimization of the objective function $\mathcal{L}(w)$ defined by:
%
    \begin{align}
    \mathcal{L}(w) &= \frac{1}{n} \sum_{i=1}^n f_i( w ) + \lambda \, \gamma( w ), \label{eq:objective}
    \end{align}
%
where $f_i(w)$ and $\gamma(w)$ are assumed to be differentiable, we consider a first-order optimization algorithm leading to the following gradient descent step at each iteration $t$:
%
    \begin{align}
    w^{t+1} \coloneqq w^{t} - \eta^{t} \left( \frac{1}{n} \sum_{i=1}^n \nabla f_{i}(w^{t}) + \lambda \nabla \gamma(w^t) \right), \label{eq:update:vanilla}
    \end{align}
%
where we denote by $\nabla f_{i}(w^{t})$ gradient of $f_i$ with respect to $w$ at iteration $t$, and by $\eta^{t}$ the learning rate at iteration $t$.  
The computation of the above full gradient over the entire training data is often intractable due to a large number of data, which leads to the use of stochastic gradient that is computed using a subset uniformly selected at random from the training data.
The iterative step of the stochastic gradient descent algorithm at iteration $t$ reads:
%
    \begin{align}
    w^{t+1} \coloneqq w^{t} - \eta^{t} \left( \frac{1}{B} \sum_{i \in \beta^t} \nabla f_{i}(w^{t}) + \lambda \nabla \gamma(w^t) \right), \label{eq:update:std}
    \end{align}
%
where $\beta^t$ denotes a mini-batch that is the index set of a subset uniformly selected at random from the training data. 
The mini-batch size $B = |\beta^t|$ is known to be related to the variance of the gradient norms, and thus to the regularization of the model. 
We assume that the mini-batch size is fixed in the optimization procedure to simplify the problem and emphasize the role of regularization parameter $\lambda$.
%
%
\section{Regularization via Adaptive Weight-Decay (AdaDecay)} \label{sec:adadecay}
We present a regularization algorithm that is designed to determine the degree of regularization for each model parameter considering its current state of solution in the course of optimization procedure in an adaptive way.
The optimization of interest aims to minimize the objective function that consists of a data fidelity term, a regularization term, and a control parameter for their relative weight.
The control parameter that determines the relative significance between the data fidelity and the regularization is generally chosen to be constant based on the assumption that the underlying distributions of the residual and the prior smoothness follow uni-modal distributions.
However, it is often ineffective to model the trade-off between the data fidelity and the regularization distributions using a static control parameter based on the ratio between the variances of their distributions.
Thus, we propose an adaptive regularization scheme that considers residual in the determination of regularity for both the spatial domain of model parameters and the temporal domain of optimization.  
%
%
%
\subsection{Weight-Decay for Individual Model Parameter} 
The computation of empirical stochastic gradient involves the noise process following a certain distribution with zero mean, and its variance is related to the degree of regularization that is desired to be imposed. 
We consider the regularization $\gamma(w)$ in Eq.~\eqref{eq:loss} by the squared Euclidean norm leading to the following objective function:
%
    \begin{align}
    \L(w) &= \rho( w ) + \frac{\lambda}{2} \, \| w \|_2^2, \label{eq:objective:l2}
    \end{align}
%
where $\lambda \in \R$ denotes the coefficient for the regularization term.
Then, the gradient descent step at each iteration $t$ by a first-order optimization algorithm reads:
%
    \begin{align}
    w^{t+1} & \coloneqq w^{t} - \eta^{t} \left( \nabla \rho( w^t ) + \lambda w^t \right) \nonumber \\ 
            & = (1 - \eta^{t} \lambda) \, w^{t} - \eta^{t} \, \nabla \rho(w^{t}), \label{eq:update:weightdecay}
    \end{align}
%
where $\eta^t \lambda$ is constrained to be $[0, 1)$ leading to the shrinkage of the unknown model parameters $w$ in iteration $t$ and this regularization scheme based on the $L_2^2$ norm is called weight-decay. 
In contrast to the static coefficient $\lambda \in \R$ in the conventional weight-decay regularization, we propose a regularization scheme that is designed to impose adaptive regularity to each model parameter $w_j$ with an additional term $\theta_j$ as follows:
%
    \begin{align}
    \L(w) &= \rho( w ) + \frac{\lambda}{2} \, \| \theta \odot w \|_2^2, \label{eq:objective:l2:parameterwise}
    \end{align}
%
where $\theta = (\theta_1, \theta_2, \cdots, \theta_m) \in \R^{m}, w = (w_1, w_2, \cdots, w_m) \in \R^{m}$ and the symbol $\odot$ denotes the Hadamard product defined by:
$\theta \odot w = (\theta_1 w_1, \theta_2 w_2, \cdots, \theta_m w_m)$.
The degree of regularization for each model parameter $w_j^t$ at iteration $t$ is determined by the adaptive term $\theta_j^t$, leading to the following modified iterative update step:
%
    \begin{align} \label{eq:update:weightdecay:adaptive}
    w_j^{t+1} & \coloneqq (1 - \eta^{t} \lambda \theta_j^{t}) \, w_j^{t} - \eta^{t} \, g_j^t,
    \end{align}
where $g_j^t = \frac{\partial \rho(w^t)}{\partial w_j}$ is the gradient of the data fidelity $\rho$ with respect to the parameter $w_j$ at iteration $t$. 
%
The weight-decay coefficient $\lambda \ge 0$ determines the constant degree of regularity for all the model parameters whereas the adaptive term $\theta$ determines the relative significant of each model parameter at each step in the course of optimization, i.e., the decay rate for each model parameter $w_j^t$ at iteration $t$ is determined by the global regularization parameter $\lambda$ multiplied by the adaptive term $\theta_j^t$.  
Note that Eq.~\eqref{eq:update:weightdecay:adaptive} with $\theta_j^t = 1$ for all $j$ and $t$ becomes the same as Eq.~\eqref{eq:update:weightdecay} with a constant weight-decay.

%
%
%
\subsection{Adaptive Weight-Decay based on Residual} 
We now consider the parameter-wise adaptive term $\theta_j^{t}$ in Eq.~\eqref{eq:update:weightdecay:adaptive}.
Our proposed regularization scheme is designed to impose an adaptive regularity to each model parameter based on its associated residual at each iteration of optimization leading to an adaptive regularization in both the spatial domain of the model parameter and the temporal domain of the optimization. 
The degree of regularity for each model parameter is determined in consideration of residual, or norm of gradient, that determines a discrepancy between the current state of model and the observation. 
The gradient norm $|g_j^t|$, known as Gauss-Southwell rule, has been used in importance sampling of parameters, e.g., ~\cite{glasmachers2013accelerated,nutini2017let,namkoong2017adaptive}.  This is, however, not directly applicable to our case since the
magnitude of gradients varies exponentially over layers in deep model.
We thus normalize the gradient norm $|g_j^t|$ to have mean $0$ and standard deviation (std) $1$ within each layer at each iteration in order to consider the relative significance of the local parameters within the layer. The normalized gradient-norm $\tilde{g}_j^t$ is given by:
%
    \begin{align}
    \tilde{g}_j^t &= \frac{|g_j^t| - \mu_l^t}{\sigma_l^t}, \label{eq:grad:normalized}
    \end{align}
%
where $l$ denotes the index of the layer that includes the parameter $w_j$, and $\mu_l^t$ and $\sigma_l^t$ denotes the mean and standard deviation of all the gradient norms for the parameters within the layer $l$ at iteration $t$, respectively.
We assume that the degree of regularity $\theta_j^t$ for each parameter $w_j$ at iteration $t$ follows a distribution of the residual leading to the following data-driven regularity:
%
    \begin{align}
    \theta_j^t \propto \tilde{g}_j^t, \label{eq:weight:distribution}
    \end{align}
%
where the degree of regularization for each parameter is proportional to the norm of its gradient.
In the determination of our adaptive regularization, we use the scaled sigmoid function defined by:
$S(x; \alpha) = 2 / (1 + \exp(- \alpha x))$, 
where $\alpha \in \R$ is a control parameter for the steepness of function value transition.
Then, the relative degree of regularization $\theta_j^t$ for each parameter $w_j$ at iteration $t$ is determined by the scaled sigmoid function $S$ of the normalized gradient norm $\tilde{g}_j^t$ as follows:
%
    \begin{align}
    \theta_j^t = S(\tilde{g}_j^t; \alpha) = \frac{2}{1 + \exp(- \alpha \tilde{g}_j^t)}, \label{eq:theta}
    \end{align}
%
where $\alpha$ determines the slope of the decay rate transition according to the gradient norm, and $\theta_j^t$ ranges from $0$ to $2$ and its average is $1$ since $\tilde{g}_j^t$ is normalized to have mean $0$ and standard deviation $1$.
\par
The pseudo code of the proposed algorithm for the adaptive weight-decay is described in Algorithm~\ref{alg:adadecay} where the degree of regularization for each parameter is determined based on the scaled sigmoid function of the norm of its gradient leading to the adaptive regularization in both the spatial domain of model parameters and the temporal domain of optimization.
The complexity of AdaDecay shown in Algorithm~\ref{alg:adadecay} remains $O(1)$ to the number of training examples as SGD with constant weight-decay.
\begin{algorithm}[htb]
\caption{\small Adaptive Weight-Decay (AdaDecay)}
\label{alg:adadecay}
\begin{algorithmic}
\small
\STATE $\{g_j^t\}_j$ : gradients of the parameters $j$ computed by back-propagation at iteration $t$
\STATE $\eta^t$ : learning rate at iteration $t$
\STATE $\lambda$ : global weight-decay coefficient
\STATE $\alpha$ : hyper-parameter for the adaptation to the gradient norm
\vspace{8pt}
            \FORALL {$l$ : index for the layer of the neural network} 
                \STATE $\mu_l^t$ : compute the mean of gradient norms $|g_j^t|$ in the layer $l$
                \STATE $\sigma_l^t$ : compute the std of gradient norms $|g_j^t|$ in the layer $l$
                \FORALL {$j$ : index for the model parameter $w_j$ in the layer $l$}
                    \STATE $\tilde{g}_j^t = (|g_j^t| - \mu_l^t) / \sigma_l^t$
                    \STATE $\theta_j^t = S(\tilde{g}_j^t ; \alpha)$ using Eq.~\eqref{eq:theta}
                    \STATE $w_j^{t+1} \coloneqq (1 - \eta^{t} \lambda \theta_j^{t}) \, w_j^{t} - \eta^{t} \, g_j^t$
                \ENDFOR
            \ENDFOR
\vspace{3pt}
\end{algorithmic}
\end{algorithm}
%
%

%
%
%

\section{Experimental Results} \label{sec:experimental}
We provide quantitative evaluation of the presented AdaDecay in comparison to the state-of-the-art optimization algorithms.
For experiments, we use three of major benchmark datasets on image recognition: MNIST, Fashon-MNIST, and CIFAR-10.
MNIST~\cite{lecun1998gradient} is a simple yet fundamental dataset that consists of 60K training and 10K test gray images of hand-written 10 digits.
Fashion-MNIST~\cite{xiao2017online} is a modern dataset that consists of 60K training and 10K test gray images with 10 categories of clothes and fashion items.
CIFAR-10~\cite{krizhevsky2009learning} is a more challenging task that consists of 50K training and 10K test object images with 10 categories.
Regarding the network architecture,
we employ four of shallow networks and four of deep networks:
The shallow networks include: 
fully-connected neural networks with two hidden layers (NN-2) and with three hidden layers (NN-3)~\cite{blum1991approximation},
LeNet-4~\cite{lecun1998gradient} with two convolution layers followed by two of fully-connected layers, and VGG-9~\cite{simonyan2014very}.
The deep networks used in our experiments are:
ResNet-18~\cite{he2016deep,he2016identity},
ResNet-50~\cite{Balduzzi2017shattered},
GoogLeNet~\cite{szegedy2015going},
and the densely connected convolutional networks (DenseConv)~\cite{huang2017densely}.
The batch normalization~\cite{ioffe2015batch} is used in
VGG-9, ResNet-18, ResNet-50, GoogLeNet and DenseConv.
\par
Our comparative analysis involves the following optimization algorithms:
the stochastic gradient descent with the constant weight-decay (SGD),
SGD with RMSprop~\cite{tieleman2012lecture} (RMS),
SGD with Adam~\cite{kingma2014adam} (Adam),
Entropy-SGD~\cite{Chaudhari2017EntropySGD} (eSGD),
Accelerated-SGD~\cite{Kidambi2018Acc} (aSGD),
and SGD with the presented adaptive weight-decay (AdaDecay).
Regarding hyper-parameters in our experiments, 
we use a practical condition including the mini-batch size of $B=128$ with the momentum of 0.9.
The weight-decay coefficient is inspected in Section~\ref{sec:wd-tuning} and fixed at $\lambda=5\times10^{-4}$ for all the algorithms.
The hyper-parameter $\alpha$ of AdaDecay that determines the adaptation to the gradient norm is inspected in Section~\ref{sec:alpha} and fixed at $\alpha=4$.
The learning-rate annealing with a sigmoid function that starts from $\eta=0.1$ and ends at $\eta=0.001$
is applied for SGD, eSGD, aSGD, and AdaDecay based on a pre-experiment result in which SGD with the sigmoid learning-rate annealing has achieved better accuracy than those with the fixed learning rate, exponential function, and staircase.
We use grid search and set 0.95 as the weighting factor in RMSprop,
0.9 and 0.999 for the first and second momentum factors in Adam.
The learning-rate scale in RMS and Adam is set as 0.001 for the shallow networks, and 0.0001 for the deep networks.
The hyper-parameters for eSGD and aSGD are also set as the recommended in the original papers~\cite{Chaudhari2017EntropySGD,Kidambi2018Acc} including the Langevin loop number of 5 for eSGD. 
\par
We perform the training process for 100 epochs, and use the maximum and the last $10\%$-epoch mean of the validation accuracy for the test data as the evaluation measures of each trial.
For quantitative comparison, we repeat the training process of the shallow networks with MNIST and Fashion-MNIST datasets for 50 independent trials, and the deep networks with CIFAR-10 dataset for 32 trials.
We consider both the maximum of the validation accuracy across epochs and trials, and the $10\%$-trimmed average of the last $10\%$-epoch accuracy over the trials.

\subsection{Selection of Weight-Decay Coefficient}
\label{sec:wd-tuning}
The weight-decay coefficient $\lambda$ determines the balance between the stochastic loss with the regularization and thus plays a critical role in both the standard weight-decay and the presented AdaDecay.
In Figure~\ref{fig:wd-tuning}, we test the coefficient ranging in $\lambda = 1\times10^{-3}, 7\times10^{-4}, 5\times10^{-4}, ... , 1\times10^{-4}$, using MNIST with NN-2, Fashion-MNIST with NN-2, and CIFAR-10 with ResNet-18 as instances.
We compare SGD using the constant weight-decay (constant) with SGD using our AdaDecay with fixed $\alpha=4$ (ours) where the two algorithms share the same weight-decay coefficient.
Figure~\ref{fig:wd-tuning} successfully demonstrates that the presented AdaDecay overtakes the constant weight-decay irrespective of the weight-decay coefficient $\lambda$ across both datasets and the model architecture.
Based on the these results and the related works~\cite{simonyan2014very,chollet2017xception}, we employ $\lambda=5\times10^{-4}$ in our experiments.
%
%
\def \fw {120pt}
\begin{figure*} [htb]
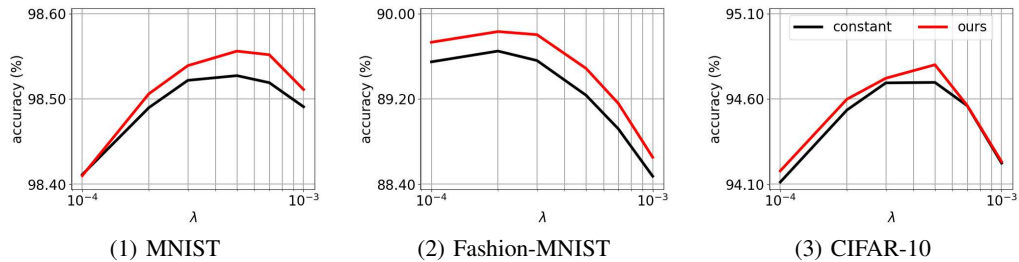

\centering
\small
\begin{tabular}{ccc}
\includegraphics[width=\fw]{{{wd.MNIST.FFNet_MNIST}}} &
\includegraphics[width=\fw]{{{wd.FashionMNIST.FFNet_MNIST}}} &
\includegraphics[width=\fw]{{{wd.cifar10.ResNet18}}} \\
(1) MNIST & (2) Fashion-MNIST & (3) CIFAR-10\\
\end{tabular}
\caption{Validation accuracy over the global weight-decay coefficient ($\lambda$) by SGD with the constant weight-decay (black) and the proposed AdaDecay with fixed $\alpha=4$ (red) for MNIST using NN-2 (left), Fashion-MNIST using NN-2 (middle), CIFAR-10 using ResNet-18 (right).
The $10\%$-trimmed average of the last $10\%$-epoch mean accuracy over 50 trials for MNIST and Fashion-MNIST, and 32 trials for CIFAR-10 is shown.
}
\label{fig:wd-tuning}
\end{figure*}

%
%
\def \pw {6.9 mm}
\def \qw {6.9 mm}
\begin{table*}[htb]
\vspace{12pt}
\caption{Validation accuracy ($\%$) of the presented AdaDecay is computed  with varying $\alpha$, or the hyper-parameter for adaptation to the gradient norm, using MNIST with NN-2 (left), Fashion-MNIST with NN-2 (middle), and CIFAR-10 with ResNet-18. The weight-decay coefficient is fixed at $\lambda=5\times10^{-4}$. 
The $10\%$-trimmed average of the last $10\%$-epoch mean accuracy (upper) and the maximum accuracy (lower) over 50 trials for MNIST and Fashion-MNIST or 32 trials for CIFAR-10 are shown.}
\label{tab.alpha.tuning}
\centering
\small
\begin{tabular}{p{6mm} | P{\qw}P{\qw}P{\qw}P{\qw}P{\pw} | P{\qw}P{\qw}P{\qw}P{\qw}P{\pw} | P{\qw}P{\qw}P{\qw}P{\qw}P{\pw}}
\hline
  & \multicolumn{5}{c|}{MNIST}  &  \multicolumn{5}{c|}{Fashion-MNIST}  &  \multicolumn{5}{c}{CIFAR-10} \\
$\alpha$ & -1 & 1 & 2 & 4 & 8 & -1 & 1 & 2 & 4 & 8 & -1 & 1 & 2 & 4 & 8\\
\hline
ave & 98.48 & 98.55 & 98.55 & 98.56 & 98.55 & 89.01 & 89.27 & 89.31 & 89.49 & 89.58 & 93.92 & 94.79 & 94.78 & 94.80 & 94.74\\
max & 98.57 & 98.67 & 98.70 & 98.72 & 98.69 & 89.35 & 89.56 & 89.68 & 89.84 & 89.92 & 94.24 & 95.05 & 94.98 & 95.04 & 94.94\\
\hline
\end{tabular}
\vspace{12pt}
\end{table*}
%

%
\subsection{Adaptation to Gradient Norm}
\label{sec:alpha}
We empirically demonstrates the effect of hyper-parameter $\alpha$ in Eq.~\eqref{eq:theta} that determines the adaptation to the gradient norm normalized in layers. 
We present the $10\%$-trimmed average and the maximum accuracy over the trials in Table~\ref{tab.alpha.tuning} where $\alpha$ is set as -1, 1, 2, 4, 8 and we trained NN-2 using MNIST and Fashion-MNIST datasets and ResNet-18 using CIFAR-10 for instances.
As shown in Table~\ref{tab.alpha.tuning}, $\alpha=-1$ is inferior to $\alpha=1$ in accuracy.  
This allows our algorithm to impose higher weight-decay rate due to larger gradient-norms.
Table~\ref{tab.alpha.tuning} also demonstrates the presented AdaDecay is robust to the choice of the hyper-parameter $\alpha>0$.
We use $\alpha=4$ throughout the following experiment, that has achieved the best result for MNIST and CIFAR10 and second best for Fashion-MNIST in Table~\ref{tab.alpha.tuning}
%
%

%
%
\subsection{Comparison to Randomized Weight-Decay}
\label{sec:comparison.to.randomized.wd}
Since we normalize the gradient norm in layer with mean of 0 and std of 1, one may argue that AdaDecay involves a randomization effect to weight-decay.
We thus compare AdaDecay with a noise injection to the weight-decay, namely randomized weight-decay, that follows Algorithm~\ref{alg:adadecay} but replaces
Eq.~\eqref{eq:theta} by 
\begin{equation} 
\theta_j^t = S(\mathcal{N}_{(0,1)}; \alpha) = \frac{2}{1 + \exp(- \alpha \mathcal{N}_{(0,1)})}, \label{eq:RND}
\end{equation}
where $\mathcal{N}_{(0,1)}$ is a random variable following the Normal distribution with mean of 0 and std of 1.
Table~\ref{tab:randomized-wd} presents the validation accuracy for MNIST, Fashion-MNIST, and CIFAR-10 by fundamental models of NN-2, NN-2, and ResNet-18 respectively, trained by SGD with constant weight-decay, the randomized weight-decay, and the proposed AdaDecay with $\lambda=5\times10^{-4}$ and $\alpha=4$.
It is successfully demonstrated that the benefit of our AdaDecay is not due to the randomization effect to the weight-decay but the use of adaptive weight-decay based on the gradient norm.
%
%
%
%
%
%
\def \pw {3.8mm} 
\def \qw {5.5mm} 
\begin{table}[htb]
\caption{Validation accuracy ($\%$) for MNIST by NN-2 (left), Fashion-MNIST by NN-2 (middle), and CIFAR-10 by ResNet-18 (right) using
SGD with the constant weight-dcay (const), the randomized weight-decay (rnd), and the presented AdaDecay (ours) with $\lambda=5\times10^{-4}$.  The rnd and ours share $\alpha=4$. 
The $10\%$-trimmed average of the last $10\%$-epoch mean accuracy (upper) and the maximum accuracy (lower) over 50 trials for MNIST and Fashion-MNIST or 32 trials for CIFAR-10 are shown.
}
\label{tab:randomized-wd}
\centering
\small
\begin{tabular}{P{4mm} | P{\pw}P{\pw}P{\qw} | P{\pw}P{\pw}P{\qw} | P{\pw}P{\pw}P{\qw}}
\hline
 & \multicolumn{3}{c|}{MNIST} & \multicolumn{3}{c|}{Fashion-MNIST} & \multicolumn{3}{c}{CIFAR-10} \\
 & const & rnd & ours & const & rnd & our & cons & rnd & ours\\
\hline
ave & 98.53 & 98.53 & \textbf{98.56} & 89.23 & 89.25 & \textbf{89.49} & 94.70 & 94.70 & \textbf{94.80}\\
max & 98.63 & 98.65 & \textbf{98.72} & 89.50 & 89.59 & \textbf{89.84} & 94.98 & 95.00 & \textbf{95.04}\\
\hline
\end{tabular}
\end{table}
%

%
%
%
\def \pw {110 pt}
\def \fw {115 pt}
\begin{figure} [htb]
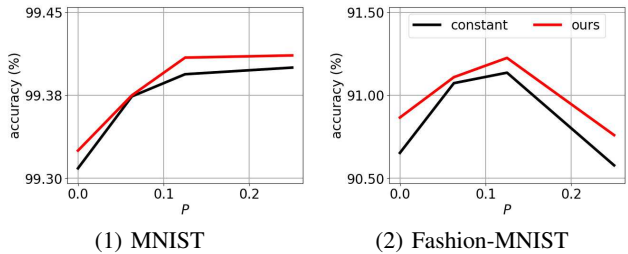

\centering
\small
\begin{tabular}{P{\pw}P{\pw}}
\includegraphics[width=\fw]{{{dropout.mnist.ConvNet_MNIST}}} &
\includegraphics[width=\fw]{{{dropout.FashionMNIST.ConvNet_MNIST}}} \\
(1) MNIST & (2) Fashion-MNIST\\
\end{tabular}
\caption{Validation accuracy over drop-out rate $P = 0, 2^{-4}, 2^{-3}, 2^{-2}$ by SGD with the constant weight-decay (black) and the proposed AdaDecay with fixed $\alpha=4$ (red) for MNIST (left) and Fashion-MNIST (right) using LeNet-4.  The weight-decay coefficient is fixed at $\lambda=5\times10^{-4}$.
The $10\%$-trimmed average of the last $10\%$-epoch mean accuracy over 50 trials for MNIST and Fashion-MNIST is shown.
}
\label{fig:dropout}
\end{figure}

%
%
\subsection{Effect of Dropout}
\label{sec:effect.of.dropout}
We demonstrate that our AdaDecay can be combined with dropout that gives an implicit regularization effect.
We present accuracy curve over the dropping rate ($P$) within 50 trials in Figure~\ref{fig:dropout} where
we trained LeNet-4 that supports the dropout for MNIST and Fashion-MNIST using SGD with the constant weight-decay of $\lambda=5\times10^{-4}$ (SGD) and with our AdaDecay (Ours) with $\lambda=5\times10^{-4}$ and $\alpha=4$.
Figure~\ref{fig:dropout} demonstrates that our AdaDecay overtakes the constant weight-decay irrespective of the use of dropout.
We employ the dropping rate of $P=0$ for LeNet-4 in the other experiment for a fair comparison with other network models.
%
%

%
%
\subsection{Comparison to State-of-the-Arts}
\label{sec:compariosn.to.soa}
We now compare SGD with the presented adaptive weight-decay (AdaDecay) to 
SGD with the constant weight-decay (SGD),
SGD with RMSprop~\cite{tieleman2012lecture} (RMS), 
SGD with Adam~\cite{kingma2014adam} (Adam), 
Entropy-SGD~\cite{Chaudhari2017EntropySGD} (eSGD),
and Accelerated-SGD~\cite{Kidambi2018Acc} (aSGD).
We fix the weight-decay coefficient at $\lambda=5\times10^{-4}$ for all the algorithms and $\alpha=4$ for ours.
In Table~\ref{tab:accuracy}, we present the $10\%$-trimmed average (upper) and the maximum (lower) validation accuracy with the shallow networks: NN-2, NN-3, LeNet-4, and VGG-9 for MNIST (1)  and Fashion-MNIST (2)  over 50 trials, and the deep models: ResNet-18, ResNet-50, GoogLeNet, and DesnseConv for CIFAR-10  (3) over 32 trials, respectively.
It is shown that SGD powered by our AdaDecay outperforms all the others in the average accuracy consistently regardless of model and dataset.
The visualization of the average accuracy curve over epochs in Figure~\ref{fig:accuracy_curve} indicating that SGD with our AdaDecay (red) achieves better accuracy than SGD with the constant weight-decay (black),
RMSprop (yellow), Adam (blue), Engropy-SGD (magenta), and Accelerated-SGD (green) across both the shallow networks with MNIST (1) and Fashion-MNIST (2), and the deep networks with CIFAR-10 (3).

%
%
\def \pw {9.7mm}
\begin{table*}[htb]
\caption{Validation accuracy ($\%$) for MNIST (top) and Fashion-MNIST (middle) by shallow models: NN-2, NN-3, LeNet-4, and VGG-9, and for CIFAR-10 (bottom) by deep models: ResNet-18, ResNet-50, GoogLeNet, and DenseConv, is computed with: SGD with the constant weight-decay (SGD), SGD with RMSprop (RMS), SGD with Adam (Adam), Entropy-SGD (eSGD), Accelerated-SGD (aSGD), and SGD with the presented AdaDecay (Ours) with $\alpha=4$. The weight-decay coeffieicnt is fixed at $\lambda=5\times10^{-4}$. The trimmed average of last $10\%$ epoch accuracy (upper part) and the maximum (lower part) over 50 trials for the shallow models and 32 trials for the deep models are shown.}
\label{tab:accuracy}
\centering
\small
(1) Validation accuracy for MNIST\\
\begin{tabular}{l | P{\pw}P{\pw}P{\pw}P{\pw}P{\pw}P{\pw} | P{\pw}P{\pw}P{\pw}P{\pw}P{\pw}P{\pw}}
\hline
 & \multicolumn{6}{c|}{NN-2}  & \multicolumn{6}{c}{NN-3}\\
 & SGD & RMS & Adam & eSGD & aSGD & Ours & SGD & RMS & Adam & eSGD & aSGD & Ours\\
\hline
ave & 98.53 & 98.22 & 98.19 & 98.16 & 98.15 & \textbf{98.56} & 98.66 & 98.31 & 98.26 & 98.29 & 98.23 & \textbf{98.69}\\
max & 98.63 & 98.36 & 98.31 & 98.31 & 98.35 & \textbf{98.72} & 98.80 & 98.49 & 98.47 & 98.41 & 98.45 & \textbf{98.82}\\
\hline
\hline
 & \multicolumn{6}{c|}{LeNet-4}  & \multicolumn{6}{c}{VGG-9}  \\
 & SGD & RMS & Adam & eSGD & aSGD & Ours & SGD & RMS & Adam & eSGD & aSGD & Ours  \\
\hline
ave &99.31 & 99.30 & 99.27 & 99.23 & 99.17 & \textbf{99.32} & 99.62 & 99.37 & 99.37 & 99.58 & 99.52 & \textbf{99.63}\\
max &\textbf{99.48} & 99.39 & 99.37 & 99.38 & 99.28 & 99.45 & \textbf{99.71} & 99.50 & 99.43 & 99.63 & 99.59 & 99.70\\
\hline
\end{tabular}\\
\vspace{6pt}
(2) Validation accuracy for Fashion-MNIST\\
\begin{tabular}{l | P{\pw}P{\pw}P{\pw}P{\pw}P{\pw}P{\pw} | P{\pw}P{\pw}P{\pw}P{\pw}P{\pw}P{\pw}}
\hline
 & \multicolumn{6}{c|}{NN-2}  & \multicolumn{6}{c}{NN-3}\\
 & SGD & RMS & Adam & eSGD & aSGD & Ours & SGD & RMS & Adam & eSGD & aSGD & Ours\\
\hline
ave & 89.23 & 88.89 & 88.98 & 87.63 & 89.12 & \textbf{89.49} & 89.71 & 89.17 & 89.26 & 88.13 & 89.24 & \textbf{89.95}\\
max & 89.50 & 89.15 & 89.28 & 87.83 & 89.44 & \textbf{89.84} & 89.87 & 89.54 & 89.49 & 88.32 & 89.56 & \textbf{90.23}\\
\hline
\hline
& \multicolumn{6}{c|}{LeNet-4}  & \multicolumn{6}{c}{VGG-9}  \\
 & SGD & RMS & Adam & eSGD & aSGD & Ours & SGD & RMS & Adam & eSGD & aSGD & Ours  \\
\hline
ave & 90.65 & 90.81 & 90.78 & 89.76 & 90.12 & \textbf{90.87} & 93.45 & 91.97 & 92.08 & 93.33 & 93.05 & \textbf{93.51}\\
max & 91.36 & 91.30 & 91.23 & 90.54 & 90.48 & \textbf{91.51} & 93.73 & 92.36 & 92.46 & 93.72 & 93.37 & \textbf{93.83}\\
\hline
\end{tabular}\\
\vspace{6pt}
(3) Validation accuracy for CIFAR-10\\
\begin{tabular}{l | P{\pw}P{\pw}P{\pw}P{\pw}P{\pw}P{\pw} | P{\pw}P{\pw}P{\pw}P{\pw}P{\pw}P{\pw}}
\hline
 & \multicolumn{6}{c|}{ResNet-18}  & \multicolumn{6}{c}{ResNet-50} \\
 & SGD & RMS & Adam & eSGD & aSGD & Ours & SGD & RMS & Adam & eSGD & aSGD & Ours \\
\hline
ave & 94.70 & 90.72 & 90.92 & 91.46 & 93.07 & \textbf{94.80} & 94.61 & 91.22 & 91.26 & 90.76 & 92.82 & \textbf{94.71}\\
max & 94.98 & 91.25 & 91.36 & 91.99 & 93.38 & \textbf{95.04} & 95.16 & 91.82 & 91.73 & 91.38 & 93.36 & \textbf{95.22}\\
\hline
\hline
 & \multicolumn{6}{c|}{GoogLeNet}  & \multicolumn{6}{c}{DenseConv} \\
 & SGD & RMS & Adam & eSGD & aSGD & Ours & SGD & RMS & Adam & eSGD & aSGD & Ours \\
\hline
ave & 94.91 & 90.48 & 90.62 & 93.00 & 93.39 & \textbf{95.17} & 94.72 & 83.17 & 83.80 & 88.05 & 90.27 & \textbf{94.91}\\
max & 95.43 & 90.94 & 90.97 & 93.45 & 93.79 & \textbf{95.50} & 95.08 & 83.60 & 84.40 & 88.60 & 90.56 & \textbf{95.22}\\
\hline
\end{tabular}
\end{table*}
%
%
%

\def \pw {100pt}
\def \fw {100pt}
\begin{figure*} [htb]
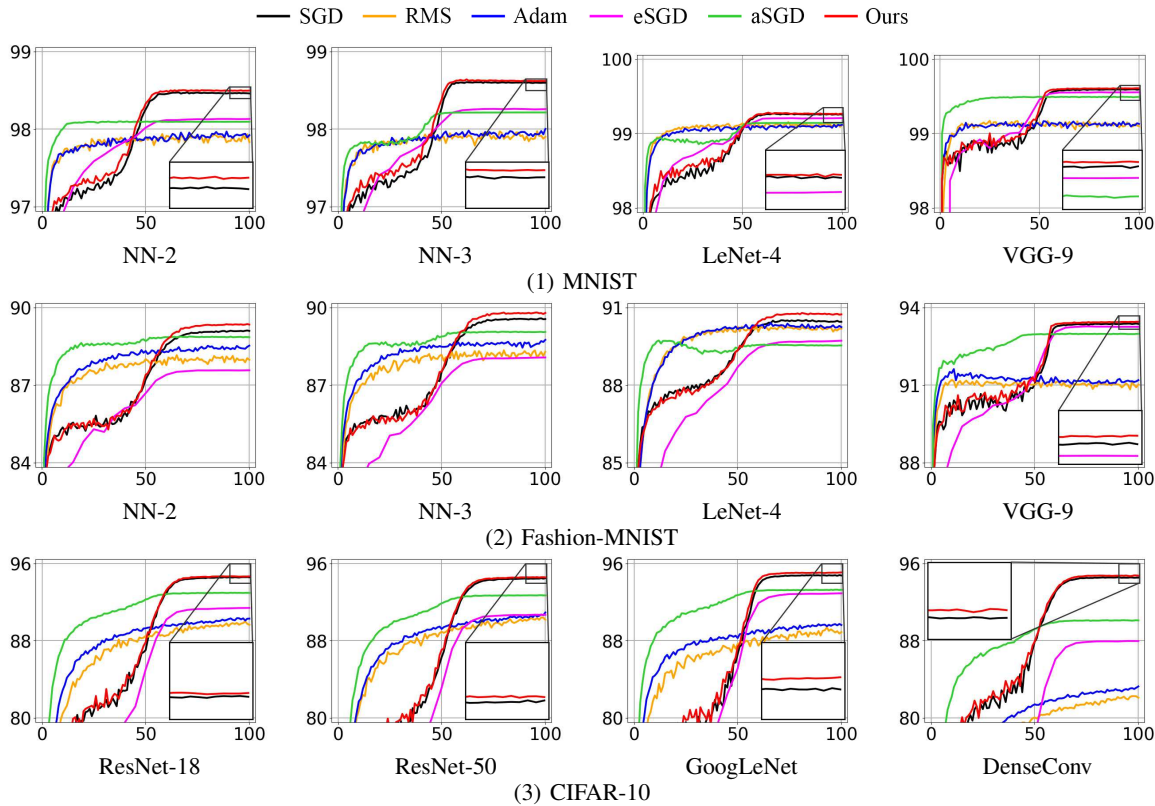

\centering
\small
\includegraphics[width=260pt]{{{legend.adadecay}}}\\
\begin{tabular}{P{\pw}P{\pw}P{\pw}P{\pw}}
\includegraphics[width=\fw]{{{cmpAccuracy,MNIST,FFNet_MNIST,e,100,B,128,mo,0.9}}} &
\includegraphics[width=\fw]{{{cmpAccuracy,MNIST,FFFNet_MNIST,e,100,B,128,mo,0.9}}} &
\includegraphics[width=\fw]{{{cmpAccuracy,MNIST,ConvNet_MNIST,e,100,B,128,mo,0.9}}} &
\includegraphics[width=\fw]{{{cmpAccuracy,MNIST,VGG9_1d,e,100,B,128,mo,0.9}}}\\
\quad NN-2 & \quad NN-3 & \quad LeNet-4 & \quad VGG-9\\
\multicolumn{4}{c}{(1) MNIST}\\
\includegraphics[width=\fw]{{{cmpAccuracy,FashionMNIST,FFNet_MNIST,e,100,B,128,mo,0.9}}} &
\includegraphics[width=\fw]{{{cmpAccuracy,FashionMNIST,FFFNet_MNIST,e,100,B,128,mo,0.9}}} &
\includegraphics[width=\fw]{{{cmpAccuracy,FashionMNIST,ConvNet_MNIST,e,100,B,128,mo,0.9}}} &
\includegraphics[width=\fw]{{{cmpAccuracy,FashionMNIST,VGG9_1d,e,100,B,128,mo,0.9}}}\\
\quad NN-2 & \quad NN-3 & \quad LeNet-4 & \quad VGG-9\\
\multicolumn{4}{c}{(2) Fashion-MNIST}\\
\includegraphics[width=\fw]{{{cmpAccuracy,cifar10,ResNet18,e,100,B,128,mo,0.9}}} &
\includegraphics[width=\fw]{{{cmpAccuracy,cifar10,ResNet50,e,100,B,128,mo,0.9}}} &
\includegraphics[width=\fw]{{{cmpAccuracy,cifar10,GoogLeNet,e,100,B,128,mo,0.9}}} &
\includegraphics[width=\fw]{{{cmpAccuracy,cifar10,DenseConv,e,100,B,128,mo,0.9}}}\\
\quad ResNet-18 & \quad ResNet-50 & \quad GoogLeNet & \quad DenseConv\\
\multicolumn{4}{c}{(3) CIFAR-10}\\
\end{tabular}
\caption{Validation accuracy curve over epoch for MNIST (top), Fashion-MNIST (middle) and CIFAR-10 (bottom) by, respectively, the shallow models (NN-2, NN-3, LeNet-4, and VGG-9) and the deep models (ResNet-18, ResNet-50, GoogLeNet, and DenseConv) optimized using the algorithms: 
SGD with constant weight-decay (black), RMSprop (yellow), Adam (blue), Entropy-SGD (magenda), Accelerated-SGD (green), and SGD with the presented AdaDecay (red) with $\alpha=4$.  The x-axis represents epoch and the y-axis represents accuracy ($\%$) at each epoch  computed as the $10\%$-trimmed average
over 50 trials for MNIST and Fashion-MNIST or 32 trials for CIFAR-10.
}
\label{fig:accuracy_curve}
\end{figure*}
%
%

%
%
%

\section{Conclusion} \label{sec:conclusion}
We have presented an adaptive regularization method for deep neural networks driven by spatio-temporal weight-decay.
The proposed algorithm is designed to consider parameter-wise weight-decay and determine it based on the norm of gradient that
reflects the current model and the given data at each optimization iteration.
The proposed AdaDecay penalizes large gradient norm and leads to better generalization of the model
independent to network architectures and is performed without any
additional cost of back-propagation or inner loop.
The robustness and effectiveness of the AdaDecay has been empirically supported by experimental results in which
SGD using our AdaDcay outperforms a number of other optimization methods for the image classification task with 
the shallow and deep networks using the major datasets.
We have focused on the image classification yet the presented adaptive regularization would have a potential impact to other machine-learning tasks using neural networks in essential.

\section*{Acknowledgment}
This work was partially supported by the National Research Foundation of Korea: 
NRF-2017R1A2B4006023 and 
NRF-2018R1A4A1059731.

\bibliographystyle{IEEEtran}
\bibliography{image}
%

%

\end{document}